\documentclass[conference]{IEEEtran}
\IEEEoverridecommandlockouts
\usepackage{cite}
\usepackage{amsmath,amssymb,amsfonts}
\usepackage{algorithmic}
\usepackage{graphicx}
\usepackage{textcomp}
\usepackage{xcolor}
\usepackage{booktabs}
\usepackage{tabularx}
\usepackage{multirow}
\usepackage{hyperref}
\hypersetup{hidelinks,
	colorlinks=true,
	allcolors=black,
	pdfstartview=Fit,
	breaklinks=true}
\def\BibTeX{{\rm B\kern-.05em{\sc i\kern-.025em b}\kern-.08em
    T\kern-.1667em\lower.7ex\hbox{E}\kern-.125emX}}
\begin{document}

\title{Continual Learning for \\Temporal-Sensitive Question Answering\\
}

\author{Wanqi Yang, Yunqiu Xu, Yanda Li, Kunze Wang, Binbin Huang, Ling Chen 
\thanks{Wanqi Yang, Yunqiu Xu, Yanda Li and Ling Chen are with University of Technology Sydney, Sydney, 2007, Australia. (email: wanqi.yang-1@student.uts.edu.au, yunqiuxu1991@gmail.com, Yanda.Li@student.uts.edu.au, ling.chen@uts.edu.au).

Kunze Wang is with University of Sydney, Sydney, 2050, Australia. (email: kwan4418@uni.sydney.edu.au).

Binbin Huang is with Hangzhou Dianzi University, Hangzhou, 310018, China. (email: huangbinbin@hdu.edu.cn).

}}

\maketitle
\begin{abstract}
In this study, we explore an emerging research area of Continual Learning for Temporal Sensitive Question Answering (CLTSQA). Previous research has primarily focused on Temporal Sensitive Question Answering (TSQA), often overlooking the unpredictable nature of future events. In real-world applications, it's crucial for models to continually acquire knowledge over time, rather than relying on a static, complete dataset. Our paper investigates strategies that enable models to adapt to the ever-evolving information landscape, thereby addressing the challenges inherent in CLTSQA. To support our research, we first create a novel dataset, divided into five subsets, designed specifically for various stages of continual learning. We then propose a training framework for CLTSQA that integrates temporal memory replay and temporal contrastive learning. Our experimental results highlight two significant insights: First, the CLTSQA task introduces unique challenges for existing models. Second, our proposed framework effectively navigates these challenges, resulting in improved performance. 

\end{abstract}


\section{Introduction}

A temporal-sensitive question refers to a question that involves temporal-related details, and modifying this temporal information within the question will result in a different answer \cite{chen2021dataset}. Take the question ``What was the role of Barack Hussein Obama in \textit{YEAR}?'' as an example. If \textit{YEAR} = 2006, the answer should be ``Federal Senator'';  whereas if \textit{YEAR} = 2016, the answer should be ``President of the United States''. 
In everyday life, we frequently encounter questions influenced by time, with answers that can change as new events occur. This unpredictability highlights the need for a novel task called Continual Learning for Temporal Sensitive Question Answering (CLTSQA), which requires continuously learn a model of temporal sensitive question answering as time progresses.

\begin{figure}[t]
\centering
\includegraphics[width=\columnwidth]{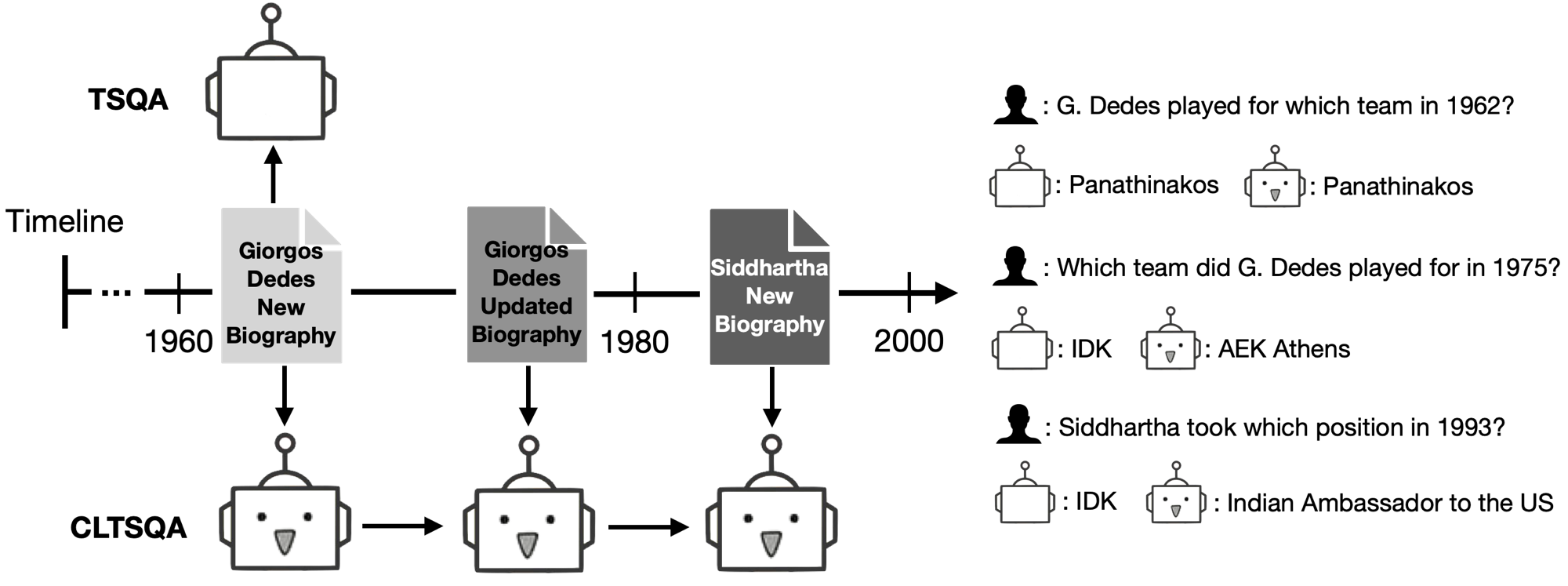}
\caption{The difference of training process between TSQA and CLTSQA. 
While TSQA assumes the availability of the whole training dataset, CLTSQA requires the model to keep ingesting up-to-date new knowledge. 
}
\label{intro}
\end{figure}

Although some works have been conducted in related areas, two key challenges of CLTSQA have been overlooked: the absence of a suitable dataset, and the scarcity of effective methods in continually dealing with temporal-sensitive questions.
While some existing works, e.g., \cite{chen2021dataset, zhang2021situatedqa, dhingra2022time, liska2022streamingqa, wang2022archivalqa}, proposed new datasets with the aim of investigating the Temporal-sensitive Question Answering (TSQA) to explore the model's sensitivity and its reasoning capabilities to temporal information. They follow the setting of traditional question answering.
As shown in Fig. \ref{intro}, TSQA assumes that the entire dataset is adapted for training the model. It lacks the ability to continuously incorporate updated and new data which could potentially alter the answer to a question as time progresses.
In terms of the second challenge, many works have been proposed to retain model's performance with evolving dataset through continual learning. For example, \cite{loureiro2022timelms} studied continual learning for a single domain (Twitter data from 2018 to 2019), and \cite{qin2022elle} worked on efficient life-long pre-training on emerging data in multiple domains. 
Currently, there are no existing efforts or studies focused on the application specifically to address CLTSQA.

The objective of the Continual Learning for Temporal Sensitive Question Answering (CLTSQA) task is to simulate a real-world scenario where updates and new knowledge cannot be learned all at once but requires continual learning. CLTSQA task explores the forgetting degree of model of knowledge in earlier time and the learning capability for acquiring updated and new knowledge over time.
To deal with the absence of an available dataset, we construct a new dataset that includes subsets of temporal-sensitive questions, thereby offering a solution to this challenge, and facilitating the study in CLTSQA.
Then, to make the model capable of effectively handling temporal-sensitive questions in a continuous fashion, we propose a novel framework featured by 1) temporal memory replay to alleviate the catastrophic forgetting of the past knowledge; and 2) temporal contrastive learning to enhance the model's sensitivity to temporal information and boost its performance on questions with most up-to-date information. 
The experimental results show that: 1) the existing models struggle to deal with this challenging task, resulting in poor performance; 2) our proposed framework can effectively help the models to address CLTSQA, demonstrating not only improvement in answering the most up-to-date questions, but also good performance retention when answering historical questions. 

The main contributions of this work are summarised as:
 \begin{itemize}
    \item We propose a novel task called CLTSQA.
    \item We propose a new dataset to deal with the absence of available dataset and facilitate the study in CLTSQA.
    \item We propose a novel framework featured by temporal memory replay and temporal contrastive learning to deal with the model-level challenge in CLTSQA.  
    \item We have obtained experimental findings indicating that: 1) CLTSQA is a challenging yet promising task, and 2) our framework assists the model in effectively addressing CLTSQA.
\end{itemize}
\section{Related Work}

\subsection{Temporal-Sensitive Question Answering}

Some previous studies have explored the task of Temporal-sensitive Question Answering by introducing new datasets. The TempQuestions dataset \cite{jia2018tempquestions} 
provides a clear definition of what constitutes a ``temporal question'' and utilizes specific trigger words such as ``before'' and ``after''. To investigate ``temporal question'', \cite{min2020ambigqa} mentioned that answers to a question can change over time and created a dataset with 13\% temporal-sensitive data.  
\cite{chen2021dataset}, \cite{zhang2021situatedqa} and \cite{wang2022archivalqa} also created new datasets, but were with a primary focus on TSQA. By evaluating existing models on the proposed datasets, these work proved that answering temporal-sensitive questions is challenging, which serves as a motivation of our study. Different from them, we not only extend TSQA towards a more realistic and challenging task CLTSQA, but also offer solutions to enhance model performance in tackling it. 

In addition to the dataset, temporal-sensitive question learning requires the model to be sensitive to temporal information. 
Several studies have utilized pre-trained language models to aid in question comprehension. However, these models do not effectively distinguish between different temporal expressions found in free-text \cite{ning2020torque,han2020econet,shang2021open,dhingra2022time}. Inspired by the framework proposed in \cite{shang2022improving}, our framework develops a temporal contrastive learning that the model can understand the crucial factor lies in recognizing the variation in temporal information, rather than the specific format of the question.

\subsection{Continual Learning}
Numerous research efforts have been dedicated to the examination of continual learning for general QA \cite{biesialska2020continual,ke2022continual}. 
Through extensive exploration of the general question answering domain, researchers have discovered that temporal-related QA tasks pose greater challenges.

\cite{liska2022streamingqa} proposed a dataset named StreamingQA, which aims to investigate models’ adaptation to changing knowledge. The dataset's context spans the years 2007 to 2020, with questions that do not involve temporally sensitive information. StreamingQA dataset employs a specific data format (question date, question, answer, document date, document), and the question date for each query is intentionally set by the author. However, datasets with additional fields and with narrower timeframes does not inherently enhance the model's robustness and generalizability.
 \cite{jang2021towards} designed a new continual learning task called continual knowledge learning (CKL). From a task-oriented perspective, the aim of CKL involves consistently enhancing the internal knowledge of the language model through ongoing pre-training on new datasets. A noteworthy distinction is that, CKL predominantly concentrates on enriching the internal knowledge within the pre-trained model, encompassing a broader domain. In contrast, CLTSQA places a stronger emphasis on a downstream task, wherein the model continuously learns and adapts to temporal-sensitive question answering.
 What's more, some temporal-related QA dataset for continual learning were proposed in \cite{loureiro2022timelms} and \cite{jang2022temporalwiki}. \cite{loureiro2022timelms} extracted data from Twitter and divided the data into subsets of three months each for continual learning. And \cite{jang2022temporalwiki} employed the difference between consecutive snapshots of English Wikipedia and English Wikidata for both training and evaluation purposes. However, they simply used the existing classical methods \cite{kirkpatrick2017overcoming,chen2020recall,wang2020k,he2021analyzing,hu2021lora} that can alleviate catastrophic forgetting in continual learning, instead of proposing improvement strategies based on their datasets.
\section{Preliminaries}
\paragraph{TSQA}

The Temporal Sensitive Question Answering (TSQA) task aims to investigate the model's sensitivity and reasoning capabilities concerning temporal information. In the TSQA, the model is provided with a context $c$ (e.g., a document, or a series of sentences) and a question $q$ as the input. Then, the model is required to predict the answer $a$ through either extracting from $c$, or selecting one from a set of answer candidates. The specific task setup for TSQA involves training the model on an entire dataset.
In order to answer temporal-sensitive questions, the model is required to not only pay specific attention to temporal information within the question, but also be capable of reasoning over the implicit temporal information within the context.

\paragraph{CLTSQA}
The TSQA task is conducted with the assumption that the model is trained using a complete dataset, However, it does not possess the capability to continuously integrate updated or new data with temporal information.
In order to alleviate this assumption, thus bridging the gap between TSQA and the real world temporal-sensitive problems, we propose a new task, CLTSQA, which forces the model to learn and inference in a continual learning manner. 
Their major difference lies in the dataset and training settings. 
Instead of assuming the availability of a whole dataset, in CLTSQA we require the model to keep awareness of the latest knowledge, while not forgetting the old knowledge. The training data is divided into $K$ subsets $\mathcal{D}=\{\mathcal{D}_1,\dots, \mathcal{D}_K\}$, with each subset covering time points that are chronologically earlier than those in the subsequent subset $t_{\mathcal{D}_{k-1}} < t_{\mathcal{D}_{k}}$.
Given an initial model $M_0$, it will be subsequently trained on the subsets to obtain the corresponding trained models $M_1, M_2, ..., M_K$, where $M_k$ denotes the model after training on $\mathcal{D}_1, \mathcal{D}_2, ..., \mathcal{D}_k$. 
The subsequent models sequentially load the pre-trained weights of the previous model and continual training. The model $M_k$ is required to be well-performing on the current dataset of $\mathcal{D}_k$, while not encountering significantly performance decay in the previous subsets $\mathcal{\overline{D}}_{k-1}$.

\section{CLTSQA Dataset}

\begin{figure*}[ht]
\centering
  \includegraphics[width=\textwidth]{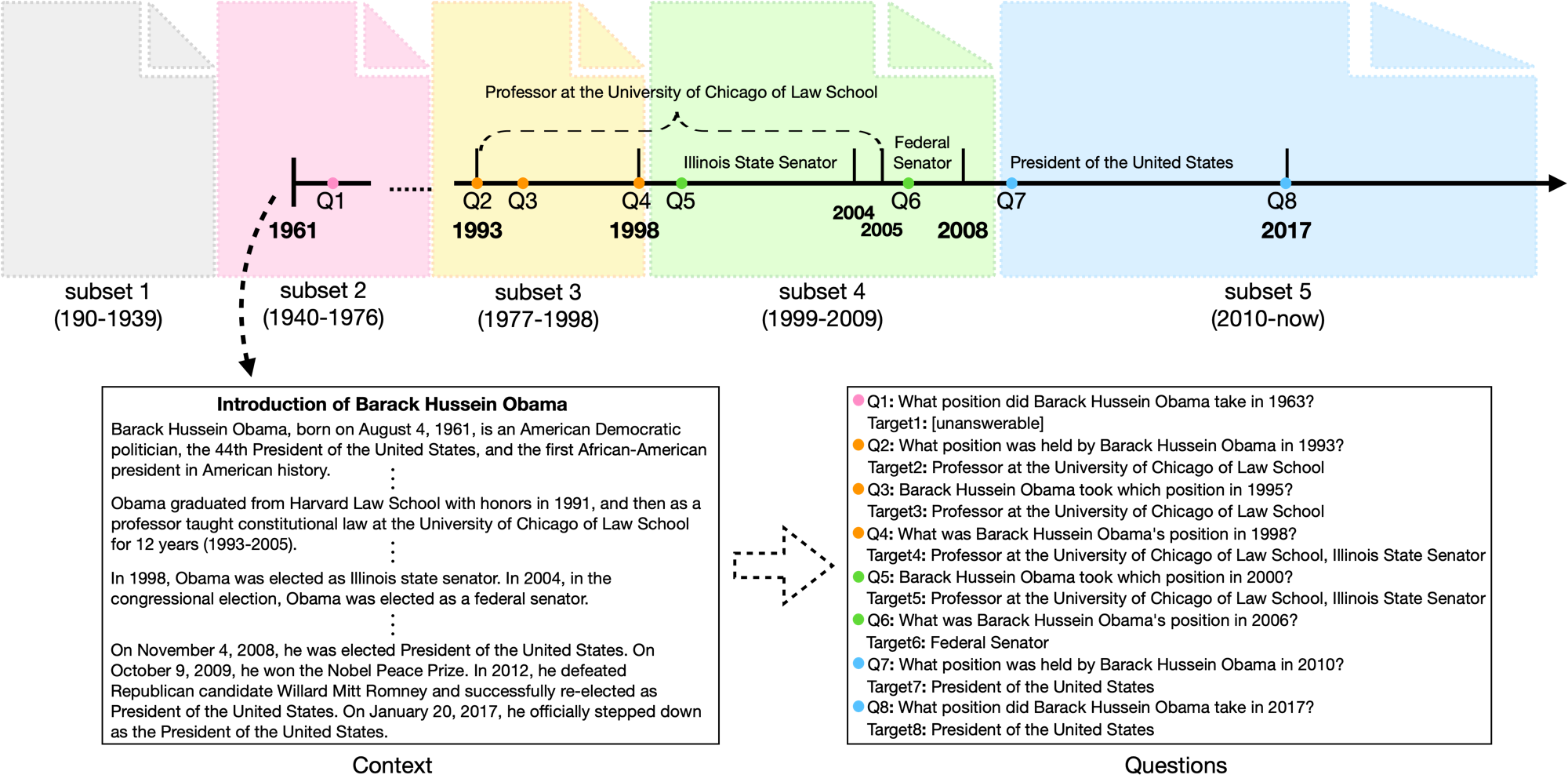}
  \caption{Examples of CLTSQA-Data. The above part shows the dataset divided based on time intervals. In the bottom part, the left side represents the context and the right side represents the corresponding question-target pairs.}
    \label{dataset}
 \end{figure*}
 
In this section, we introduce a new dataset - CLTSQA-Data, with the aim of addressing the aforementioned data-level challenge. Our dataset is built on the basis of TimeQA~\cite{chen2021dataset}, which extracts time-evolving contexts from WikiData, and generates question-answer pairs from these contexts by some manual templates. 

We chose a collection of 20,000 questions and 5,000 contexts sourced from TimeQA. Moreover, we produced a higher volume of context-specific temporal-sensitive questions. As a result, our dataset now encompasses a total of 50,000 questions and 5,000 contexts. Then we divides the whole dataset into $K$ temporal-sensitive subsets $\mathcal{D}= \{\mathcal{D}_{1}, \mathcal{D}_{2},\dots, \mathcal{D}_{K}\}$. 
Fig.~\ref{dataset} shows some examples, where each subset $\mathcal{D}_{k}$ consists of questions within a specific time range $[ t_k^{start}, t_k^{end} ]$. 
We keep the original context unchanged and generate questions based on it, then assign them to subsets with non-overlapping time ranges. For example, given a long context ``Introduction of Barack Hussein Obama'', which ranges from 1961 to 2017, we generate a series of related questions, such as 
``What position did Barack Hussein Obama take in 1963?'', ``What position was held by Barack Hussein Obama in 1995?'', ``Barack Hussein Obama took which position in 2010?'', then put them into different subsets based on time periods. Besides the explicit questions, whose answers could be directly extracted from the context, we also generate the more challenging implicit questions, whose answers could not be directly obtained, and require the model to reason from the implicit temporal relation. For example, given the context ``Barack Hussein Obama won re-election in the 2012 presidential election'', the answer to the question ``Who is the President of the United States in 2014'' should be ``Barack Hussein Obama''.

\begin{table}[ht]
\caption{\label{dataset-statistics}
The statistics of CLTSQA-Data divided by subsets \& question types. 
}
\centering
\resizebox{0.95\columnwidth}{!}
{
\begin{tabular}{l|ccc}
\toprule
\textbf{ } & \textbf{Train} & \textbf{Dev} & \textbf{Test}\\
\midrule
{Subset1 (190-1939)} & {7091} & {1562} & {1455} \\
{Subset2 (1940-1976)} & {6957} & {1405} & {1531} \\
{Subset3 (1977-1998)} & {6962} & {1493} & {1494} \\
{Subset4 (1999-2009)} & {7216} & {1415} & {1584} \\
{Subset5 (2010-now)} & {6788} & {1549} & {1344}\\
\midrule
{Easy Reasoning} & {4068} & {909} & {880} \\
{Common Sense} & {3252} & {730} & {728}\\
{Multi-descriptions Join} & {6128} & {1412} & {1260} \\
{Multi-paragraphs Join} & {15265} & {3097} & {3211} \\
{Unanswerable} & {6301} & {1276} & {1329} \\
\midrule
\textbf{Total} & {35014} & {7424} & {7408} \\
\bottomrule
\end{tabular}
}

\end{table}

\begin{figure*}[ht]
\centering
  \includegraphics[width=0.95\textwidth]{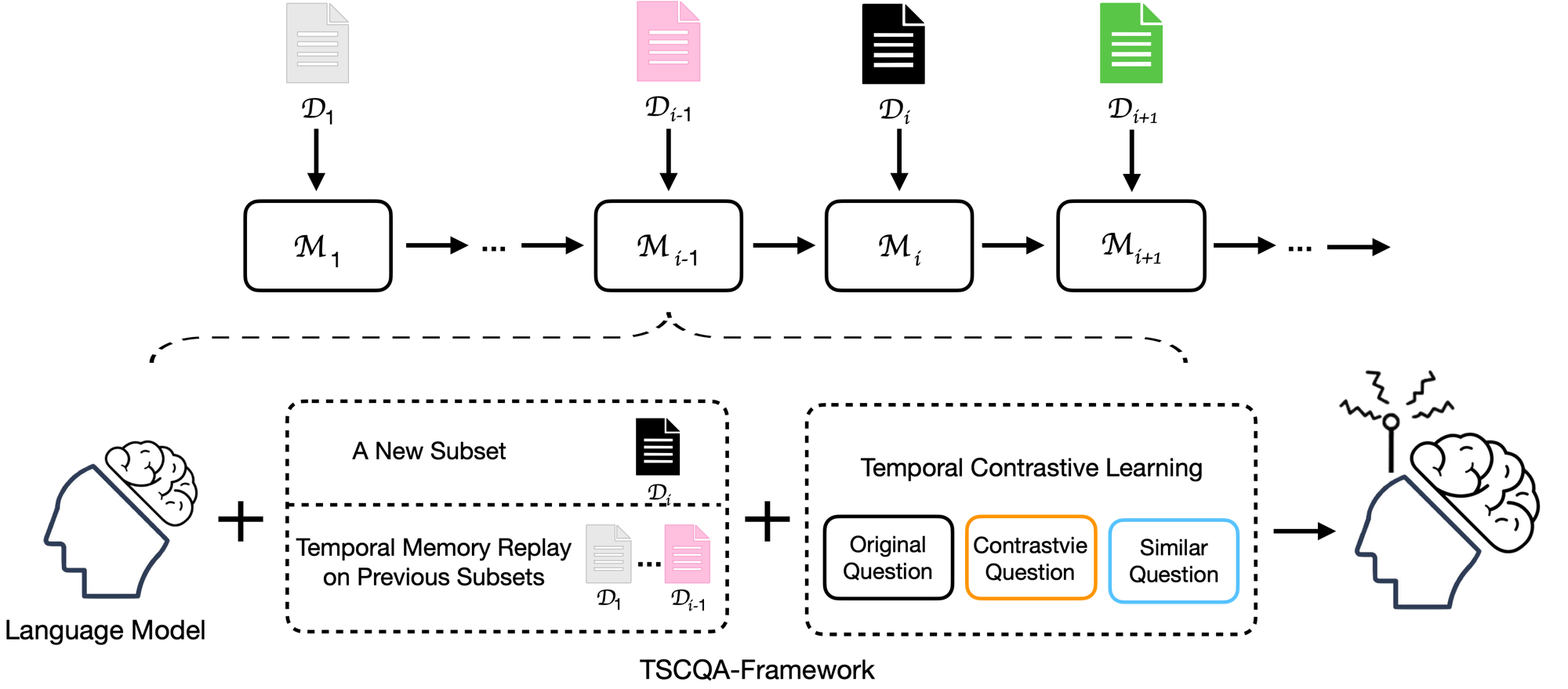}
  \caption{An overview of the CLTSQA task with our framework. The above figure illustrates the sequential learning of different subsets. The below figure represents our approach of loading the pre-trained weights of the previous model for the next model, while incorporating temporal memory replay and temporal contrastive learning.}
    \label{model}
 \end{figure*}

Table~\ref{dataset-statistics} shows the statistics the CLTSQA-Data dataset. Our dataset contains a total of 50,000 questions and 5,000 contexts. We construct $K=5$ subsets, which are made of varying time spans to ensure that they have similar amount of data. The questions could be divided into 5 types:
\begin{itemize}    
    \item \textbf{Easy reasoning}, where the temporal information in the question is explicitly specified in the context.     
    \item \textbf{Joining commonsense}, which requires the model to understand the temporal commonsense knowledge. Such as 2010 is included within 2008-2017.    
    \item \textbf{Joining multiple descriptions}, which requires the model to reason the context from multiple descriptions within the same paragraph.     
    \item \textbf{Joining multiple paragraphs}, which is a multi-paragraph extension of \textbf{Joining multiple descriptions} - the model is required to reason the context across multiple paragraphs. Joining multiple paragraphs not only limits to adjoining paragraphs, but it also extends to cases where significant temporal gaps exist between paragraphs that must be integrated. For the introductory passage about Giorgos Dedes, where the initial paragraph delineates his birth year as 1943, followed by subsequent paragraphs narrating his life at ages 30 and 40. Failing to incorporate contextual information from earlier periods would render it challenging to address inquiries such as ``Which team did Giorgos Dedes play for in 1973/1983?''. This underscores the importance of seamlessly weaving old and new text and the importance of continuous learning.
    \item \textbf{Unanswerable}, where the answer could not be found or reasoned from the context. 
    According to the description in a context, ``Barack Hussein Obama was born in August 1961'', we cannot answer the question ``What position did Barack Hussein Obama hold in 1960?''. 
\end{itemize}

\begin{figure*}[ht]
\centering
  \includegraphics[width=0.95\textwidth]{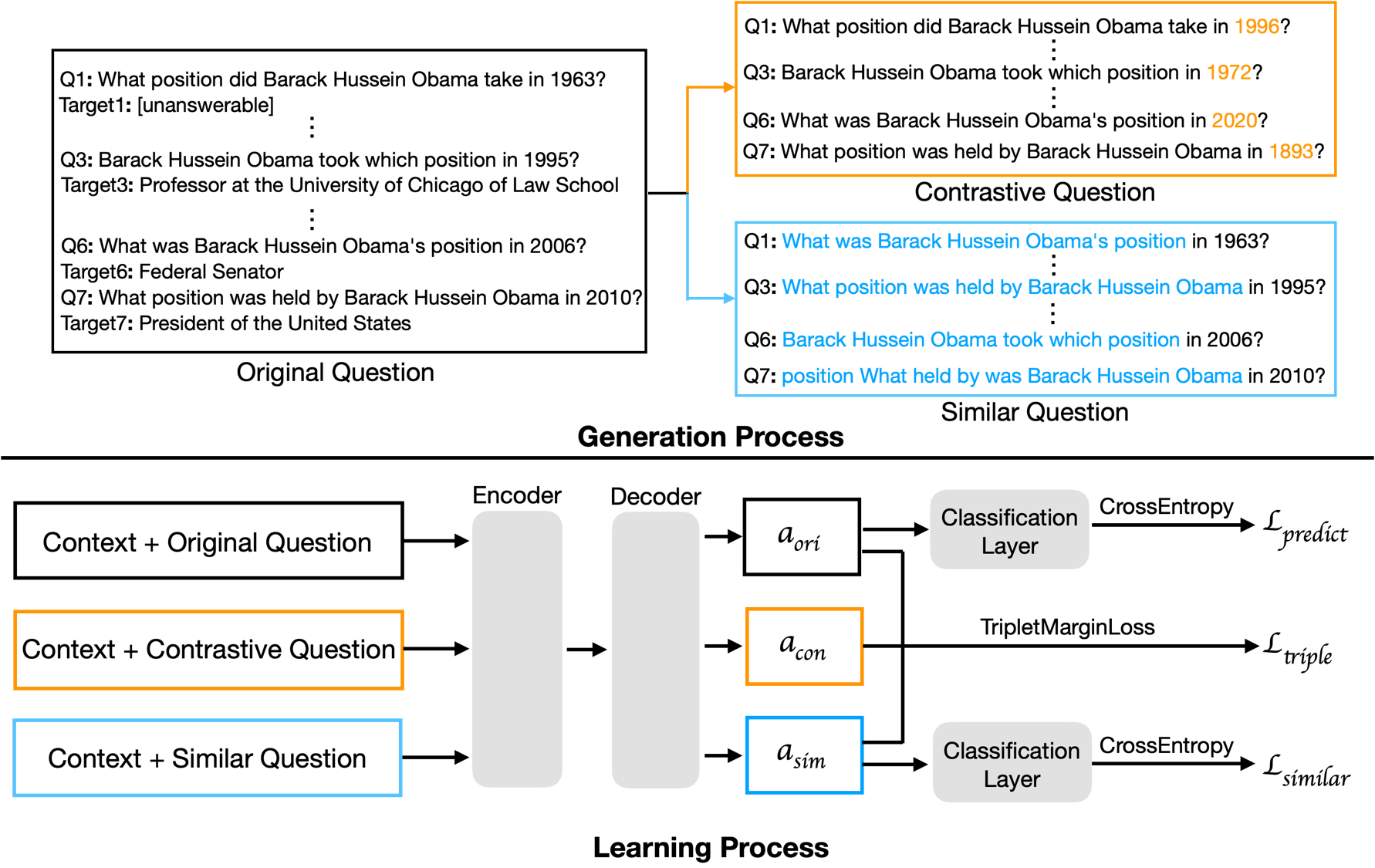}
  \caption{Illustration of temporal contrastive learning, including generation process of contrastive and similar questions as well as model's learning process.}
    \label{temporal contrastive learning}
 \end{figure*}

\section{CLTSQA Framework} \label{sec_framework}
 
In this section, we propose a model-agnostic framework - CLTSQA-Framework to address the aforementioned model-level challenge, thus helping an arbitrary model to learn the CLTSQA task. 
Fig.~\ref{model} gives an overview of our framework, which consists of two key features 1) \textbf{temporal memory replay}, and 2) \textbf{temporal contrastive learning}. 

Initialized with a pre-trained language model $M_0$, we follow the task setting in the Preliminaries section to sequentially train the model on different subsets, where $M_i$ denotes the model after training $M_{i-1}$ on the subset $\mathcal{\overline{D}}_{i-1}$. 
The first key feature is \textbf{temporal memory replay}, which inherits from continual learning to alleviate the forgetting problem during training on the new subset. Specifically, a portion of the data from the time period preceding the new subset is stored, and then replayed during the learning process of the new subset.
The second key feature is \textbf{temporal contrastive learning}, which aims at enhancing model's sensitivity to the temporal information within the questions. 
Specifically, it involves creating two additional questions based on the original question, and then combining a context along with these questions as three separate inputs for the model.

\subsection{Temporal Memory Replay}
\label{sec:TMR}

One of the key properties of the CLTSQA task, is the continual learning process, which is always accompanied by the catastrophic forgetting problem - the model tends to ``forget'' the old knowledge during ingesting the new knowledge \cite{mccloskey1989catastrophic}. 
For the temporal-sensitive questions, in particular, after acquiring knowledge about a new question, which shares a similar context to an old question except for the temporal information, the model might encounter difficulties when re-trying to answer the old question. 
For example, the model might get in trouble in answering ``Who is the president of United States in 2009'' after learning the new knowledge about ``Who is the president of United States after 2020?''. 
Motivated by the memory replay \cite{rebuffi2017icarl}, which helps the model to remember old knowledge through retaining some old training data and reusing them in the subsequent training process, we propose a temporal memory replay strategy that is for dealing with catastrophic forgetting of the data from the previous time periods.
Specifically, as the choice of which data to retain plays a crucial role in temporal memory replay, we aim to prioritize the model's attention towards data that are 1) easily learnable samples for efficiently keeping previous knowledge and 2) susceptible to distraction within the new dataset.

Take the model $M_{i-1}$ as an example, which has been sequentially trained on the previous subsets $\mathcal{\overline{D}}_{i-1}$, and will be trained on the current subset $\mathcal{D}_{i}$.
1) To better retain data from previous time periods, we removed the top $\mu$ of the hardest samples from the preceding subsets $\mathcal{\overline{D}}_{i-1}$, while retaining the easily learnable ones. This approach mitigates the challenge of data forgetting. Notably, the term ``hard sample'' is used to describe the sample that received the lowest evaluation score among the previous subsets.
2) From a temporal perspective, we select a part ($\nu$) of data from previous time periods that had the same context but different answers, and incorporated them into the new subset. By introducing these distractors, we aimed to enhance the model's robustness and its sensitivity for temporal information.


\subsection{Temporal Contrastive Learning}

CLTSQA-Data generates multiple questions based on a single context, where the questions have identical content but vary in their temporal information and expression. To enhance the model's sensitivity to temporal information in questions and acknowledge that differences in question expression do not affect the answer, the strategy of temporal contrastive learning is employed. Fig.~\ref{temporal contrastive learning} shows the strategy encompassing the generation procedure for contrasting and similar questions, along with the learning process employed by the model.

\paragraph{Generation of Contrastive and Similar Question.}
We generate a contrastive question $q_{contrast}$ and a similar question $q_{similar}$ for the original question $q$ of each sample in the training dataset. 

To create the contrastive question $q_{contrast}$, we simply substitute the temporal information in the original question with different temporal references while keeping everything else unchanged. For example, the contrastive question of the original question ``What position did Barack Hussein Obama hold in 2010?'' is ``What position did Barack Hussein Obama hold in 1995?''. It should be emphasized that the answer to the contrastive question consistently differs from the answer to the original question, thereby ensuring their distinctiveness.

To generate a similar question $q_{similar}$, we maintain the temporal information while modifying the wording of the question.
If there are alternative expressions of the original question available in CLTSQA-Data dataset $\mathcal{D}$, then substitute the expression of the original question with one of those alternatives. The original question ``What position did Barack Hussein Obama hold in 2010?'' can be transformed to a similar question ``Barack Hussein Obama took which position in 2010?''. If no other expression exists in CLTSQA-Data dataset, We process the question with word segmentation and randomly rearrange the positions of the tokens in the question, excluding the temporal information. For example, the original question is ``What position did Barack Hussein Obama hold in 2010?'', and its similar question is ``position What Barack Hussein Obama did hold in 2010?''. 
The study conducted by \cite{sinha2020unnatural} and  \cite{sinha2021masked} demonstrate that word order does not have a significant impact on model performance across various downstream tasks, including Question Answering (QA). Therefore, we employ the aforementioned approach to strive for consistency between similar questions and the original question.

\paragraph{Temporal Contrastive Learning.}
As Fig.~\ref{temporal contrastive learning} shows, we concatenate a context $c$ and original question ${q}_{ori}$, contrastive question ${q}_{con}$, similar question ${q}_{sim}$ respectively as the three inputs $\mathbf{x}=\{{q}_{ori}, {c}\}$, $\mathbf{x}_{con}=\{{q}_{con}, {c}\}$ and $\mathbf{x}_{sim}=\{{q}_{sim}, {c}\}$ of the model. These inputs are passed through model, obtaining three representations ${a}_{ori}$, ${a}_{con}$ and ${a}_{sim}$.

We first apply TripletMarginLoss \cite{balntas2016learning} function over $a_{ori}$, $a_{con}$ and $a_{sim}$ to obtain $L_{triple}$.
\begin{equation}
    T(s,p,n) = max\{d(s_i, p_i)-d(s_i, n_i)+margin, 0 \}
\end{equation}
where
\begin{equation}
    d(x,y) = \parallel x-y \parallel_p
\end{equation}
and
\begin{equation}
    L_{triple} = T(a_{ori},a_{sim},a_{con})
\end{equation}
Then ${a}_{ori}$ and ${a}_{sim}$ are processed by a linear layer to obtain representations $\hat{a}_{ori}$ and $\hat{a}_{sim}$. 
We get answer prediction loss $L_{predict}$ by applying CrossEntropy function over target label $a_{target}$ and the representation $\hat{a}_{ori}$. Likewise, get similar loss $L_{similar}$ by applying CrossEntropy function over target label $a_{target}$ and the representation $\hat{a}_{sim}$.

Finally we combine $L_{predict}$, $L_{similar}$ and $L_{triple}$ as the final objective function loss:
\begin{equation}
    Loss = \alpha L_{predict}+\beta L_{similar}+\gamma L_{triple}
\end{equation}
where $\alpha>0, \beta>0, \gamma>0$ are weight factors.
\begin{table*}[t]
\caption{\label{results}
Results of models' final performance after sequentially training on the 5 subsets. ``FiD-CLTSQA'' (``BigBird-CLTSQA'') and ``FiD-baseline'' (``BigBird-baseline'') denote the model trained with / without the proposed CLTSQA-Framework, respectively.
}
\centering
\resizebox{\linewidth}{!}{
\begin{tabular}{l|c|c|c|c|c|c|c|c|c|c|c|c|c|c|c|c|c|c|c|c}
\toprule
\multirow{3}*{ } & \multicolumn{4}{c|}{Subset1} & \multicolumn{4}{c|}{Subset2} & \multicolumn{4}{c|}{Subset3} & \multicolumn{4}{c|}{Subset4} & \multicolumn{4}{c}{Subset5}\\ 
\cline{2-21}
 & \multicolumn{2}{c|}{Dev} & \multicolumn{2}{c|}{Test} &  \multicolumn{2}{c|}{Dev} & \multicolumn{2}{c|}{Test}& \multicolumn{2}{c|}{Dev} & \multicolumn{2}{c|}{Test}& \multicolumn{2}{c|}{Dev} & \multicolumn{2}{c|}{Test}& \multicolumn{2}{c|}{Dev} & \multicolumn{2}{c}{Test} \\
\cline{2-21}
 &EM&F1&EM&F1&EM &F1&EM &F1&EM &F1&EM &F1&EM &F1&EM &F1&EM &F1&EM &F1\\
\midrule
{FiD-Baseline} &  34.25 & 45.29 & 29.97 & 40.22 & 43.84 & 53.89 & 39.26 & 50.95& 40.32 & 50.60 & 40.43& 49.48& 42.26 & 53.37 & 46.28& 56.60& 47.97 & 55.60 & 49.03& 56.29 \\
{FiD-CLTSQA} & \textbf{42.45} & \textbf{52.01} & \textbf{39.31} & \textbf{49.55}& \textbf{47.97} & \textbf{57.95} &\textbf{47.49} &\textbf{57.76} & \textbf{47.96} & \textbf{57.05} &\textbf{48.06} &\textbf{56.84} & \textbf{46.08} & \textbf{55.95} & \textbf{49.12}&\textbf{59.16} & \textbf{49.71} & \textbf{57.48} &\textbf{49.03} &\textbf{57.06} \\
\midrule
{BigBird-Baseline} & 31.24 & 40.55 & 29.48 &38.81 & 35.16 & 45.14 & 35.66&44.68 &26.59 &36.04 & 32.46 & 40.85& 35.76 & 43.58 & 37.94&46.48 & 41.58 & 48.08 & 41.74 &50.16 \\
{BigBird-CLTSQA} & \textbf{35.21} & \textbf{43.54} & \textbf{33.81}& \textbf{41.49}& \textbf{42.63}&  \textbf{51.57}&\textbf{42.91} & \textbf{50.64}& \textbf{38.25} & \textbf{45.53} & \textbf{42.24} & \textbf{50.22}&\textbf{39.93} & \textbf{47.90}& \textbf{39.02}& \textbf{46.83}&\textbf{43.77} &\textbf{49.72} & \textbf{44.72}& \textbf{50.42} \\
\bottomrule
\end{tabular} 
}

\end{table*}

\begin{table*}[t]
\caption{\label{Ablation}
Ablation results of model variants after sequentially training on the 5 subsets.
``TMR'' and ``TCL'' denote ``Temporal Memory Replay'' and ``Temporal Contrastive Learning'', respectively.
}
\centering
\resizebox{\linewidth}{!}{
\begin{tabular}{l|c|c|c|c|c|c|c|c|c|c|c|c|c|c|c|c|c|c|c|c}
\toprule
\multirow{3}*{ } & \multicolumn{4}{c|}{Subset1} & \multicolumn{4}{c|}{Subset2} & \multicolumn{4}{c|}{Subset3} & \multicolumn{4}{c|}{Subset4} & \multicolumn{4}{c}{Subset5}\\ 
\cline{2-21}
 & \multicolumn{2}{c|}{Dev} & \multicolumn{2}{c|}{Test} &  \multicolumn{2}{c|}{Dev} & \multicolumn{2}{c|}{Test}& \multicolumn{2}{c|}{Dev} & \multicolumn{2}{c|}{Test}& \multicolumn{2}{c|}{Dev} & \multicolumn{2}{c|}{Test}& \multicolumn{2}{c|}{Dev} & \multicolumn{2}{c}{Test} \\
\cline{2-21}
 &EM&F1&EM&F1&EM &F1&EM &F1&EM &F1&EM &F1&EM &F1&EM &F1&EM &F1&EM &F1\\
\midrule
{FiD-CLTSQA} & \textbf{42.45} & 52.01 & \textbf{39.31} & \textbf{49.55}& \textbf{47.97} & \textbf{57.95} &\textbf{47.49} &\textbf{57.76} & \textbf{47.96} & \textbf{57.05} &48.06 &56.84 & \textbf{46.08} & \textbf{55.95} & \textbf{49.12}&\textbf{59.16} & \textbf{49.71} & \textbf{57.48} &49.03 &57.06 \\
{w/o TCL} & 42.06 & \textbf{52.44} &38.63 &49.12 & 45.84 & 55.63 &45.07 &55.49 & 45.41 & 55.68 &\textbf{48.13} &\textbf{57.22} & 44.73 & 54.81 &46.28 &56.93 & 48.42 & 55.72 &47.32 &54.81  \\
{w/o TMR} & 15.94 & 19.09 & 17.59 & 20.99& 17.86 & 21.53 &19.33 &23.43 & 17.95 & 22.82 & 18.94& 22.14& 42.83 & 52.65 &43.81 &53.40 & 48.93 & 56.91 &\textbf{49.48} &\textbf{57.40} \\
{FiD-Baseline} &  34.25 & 45.29 & 29.97 & 40.22 & 43.84 & 53.89 & 39.26 & 50.95& 40.32 & 50.60 & 40.43& 49.48& 42.26 & 53.37 & 46.28& 56.60& 47.97 & 55.60 & 49.03& 56.29 \\
\bottomrule
\end{tabular} 
}

\end{table*}

\begin{table*}[t]
\caption{\label{Ablation2}
Ablation results of model variants after sequentially training on the 5 subsets.
``MR'' and ``TMR'' denote ``Memory Replay'' and ``Temporal Memory Replay'', respectively.
}
\centering
\resizebox{\linewidth}{!}{
\begin{tabular}{l|c|c|c|c|c|c|c|c|c|c}
\toprule
\multirow{3}*{ } & \multicolumn{2}{c|}{Subset1} & \multicolumn{2}{c|}{Subset2} & \multicolumn{2}{c|}{Subset3} & \multicolumn{2}{c|}{Subset4} & \multicolumn{2}{c}{Subset5}\\ 
\cline{2-11}
 &EM&F1&EM&F1&EM &F1&EM &F1&EM &F1\\
\midrule
{FiD-Baseline with MR} & 40.91 & 51.51 & 54.62 & \textbf{56.46}& \textbf{45.75} & 55.28 &43.46 &53.71 & 47.19 & 54.92 \\
{FiD-Baseline with TMR} & \textbf{42.06} & \textbf{52.44} &\textbf{45.84} &55.63 & 45.41 & \textbf{55.68} &\textbf{44.73} &\textbf{54.81} & \textbf{48.42} & \textbf{55.72} \\
\bottomrule
\end{tabular} 
}

\end{table*}

\section{Experiments}

In this section, we conduct experiments for the CLTSQA task, and would like to answer the following three research questions: 
1) whether the novel task CLTSQA poses new challenges to the existing QA models; 
2) whether our framework helps the models to deal with the CLTSQA task; and 3) which part of our framework contributes more to the performance improvement.

\paragraph{Data} We conduct the experiment upon the proposed CLTSQA-Data dataset. Specifically, we use $K=5$ subsets, each of which consists of around 7,000 training questions, 1,500 validation questions and 1,500 testing questions. Table~\ref{dataset-statistics} shows the statistics of the subsets. 

\paragraph{Model} As illustrated in Sec. \ref{sec_framework}, our framework is model-agnostic and can be applied to arbitrary QA models. We use the following two models as our baselines:
\begin{itemize}   
    \item FiD~\cite{izacard2020leveraging}, whose objective is to generate answers sequentially, token by token, in an auto-regressive manner. It has achieved impressive performance on Natural Questions \cite{kwiatkowski2019natural} and TriviaQA \cite{joshi2017triviaqa}.
    \item BigBird~\cite{zaheer2020big}, which introduces a sparse attention mechanism that enhances performance across various tasks involving extensive contextual information. This model focuses on extracting the answers from a given sequence and has achieved remarkable outcomes in question answering.
\end{itemize}

\paragraph{Training} We follow \cite{izacard2020leveraging} and \cite{zaheer2020big} to construct FiD and BigBird, and initialize the baselines with Natural Question pre-trained weights. 
For temporal memory replay, we set $\mu=10\%$ and $\nu=10\%$. For temporal contrastive learning, we set $\alpha:\beta:\gamma=1:0.5:0.5$. During training, we continuously train the model on the 5 subsets. For each subset, we train the model for 8 epochs with a batch size of 1. The model is optimized using AdamW \cite{loshchilov2017decoupled} with a learning rate of $5e^{-5}$. 

\paragraph{Evaluation} After training on a subset, we evaluate the model on the testing set of this subset as well as all previous subsets. We use exact match (EM) and F1 score as the evaluation metrics.

\section{Results and Discussions}

\subsection{Main Results}

Table~\ref{results} shows models' evaluation performance after subsequently training on the five subsets. ``FiD-CLTSQA'' (``BigBird-CLTSQA'') and ``FiD-baseline'' (``BigBird-baseline'') denote the model trained with / without the proposed CLTSQA-Framework, respectively. The baselines (``FiD-Baseline'' and ``BigBird-Baseline''), which are trained in a sequential manner but without utilizing the proposed framework (i.e., no temporal memory replay or temporal contrastive learning), exhibit poor performance. In particular, the baselines perform worst when being evaluated on Subset1, which has the greatest temporal difference from the most up-to-date subset (Subset5). Such observations answer our first research question - the current QA models may face challenges when tackling the CLTSQA task. 

When it comes to the proposed CLTSQA-Framework, it is evident that this framework helps the models to obtain improved performance, especially in those ``earlier'' subsets. Taking the earliest subset, Subset1, as an example, when equipped with CLTSQA-Framework, the BigBird model demonstrates a 14.69\% increase in EM and 6.91\% increase in F1 (``BigBird-CLTSQA'' v.s., ``BigBird-Baseline''). More significant performance improvement could be observed in FiD, which demonstrates a 31.16\% increase in EM and 23.20\% increase in F1 (``FiD-CLTSQA'' v.s., ``FiD-Baseline'').
Such observations answer our second research question - the proposed framework helps the models to deal with the CLTSQA task.

\begin{figure}[t]

\centering
\includegraphics[width=\columnwidth]{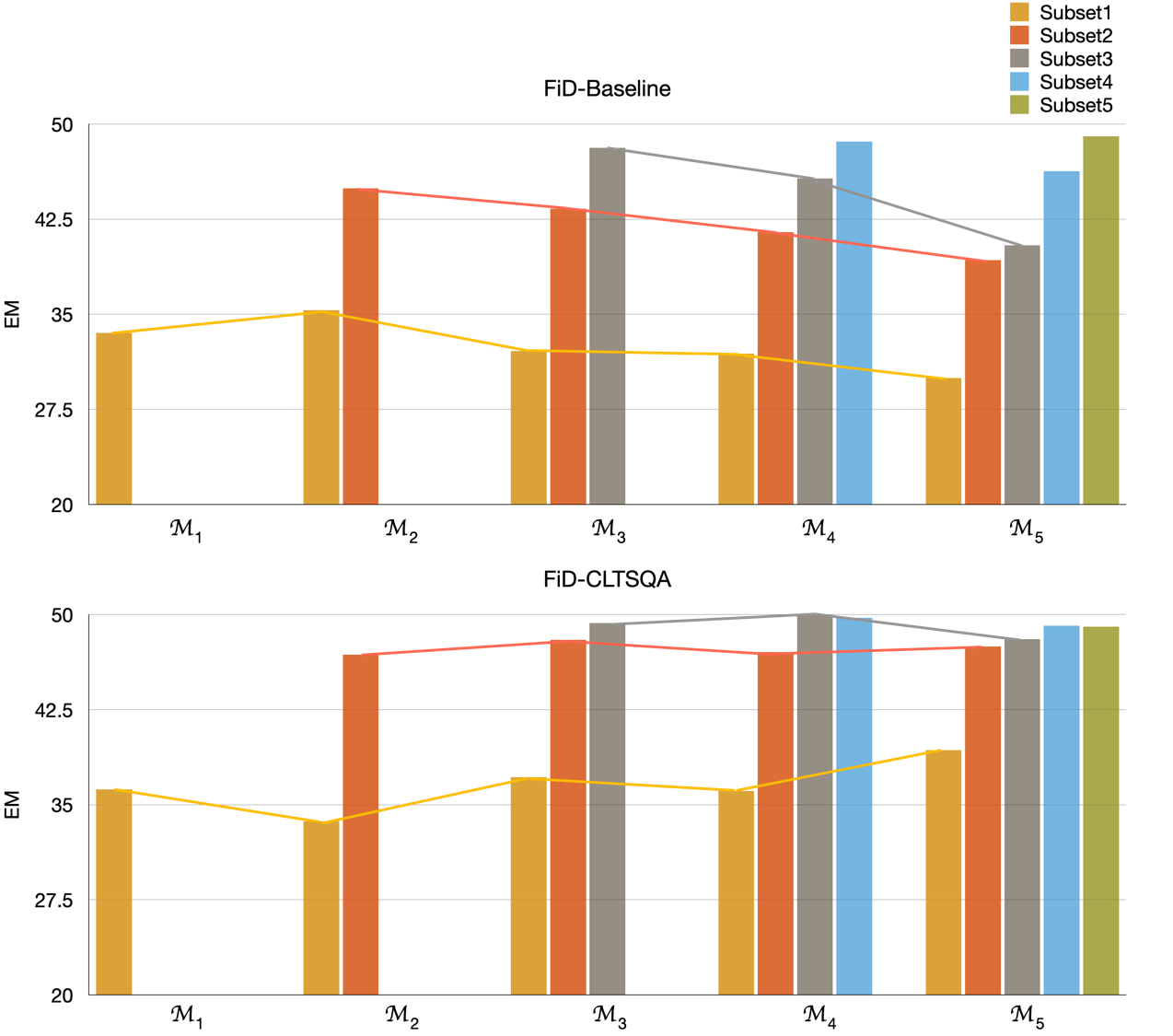}
\caption{The testing performance of ``FiD-Baseline'' and ``FiD-CLTSQA'' models in different training stages. The proposed framework effectively helps the FiD model to retain the performance on old subsets throughout the training process. }
\label{bar}
\vspace{-3mm}
\end{figure}

The significant performance improvement could be attributed to two strategies introduced by the proposed CLTSQA-Framework: 1) the temporal memory replay, which helps the model to retain the old knowledge when ingesting the latest knowledge; and 2) the temporal contrastive learning, which helps the model to acquire representations in a manner that captures and distinguishes the temporal information present in the question, thus enhancing model's ability in answering the temporal-sensitive questions. 
To validate these strategies, Fig. ~\ref{bar} shows the testing performance of ``FiD-Baseline'' and ``FiD-CLTSQA'' models in different training stages, where $M_i$ denotes the model after training on subset $\mathcal{\overline{D}}_{i}$. It could be observed that while ``FiD-Baseline'' encounters performance drop in Subset 1, Subset 2 and Subset 3 with the progress of training, ``FiD-CLTSQA'' retains its performance on those subsets throughout the training process, validating the first strategy. 
The second strategy could be validated from two perspectives. Firstly, going beyond retaining the performance, the model with CLTSQA-Framework can even improve performance on Subset 1 with the progress of training, showing the enhancement of ability of answering temporal-sensitive questions. Secondly, in the up-to-date subsets such as Subset 4 and Subset 5, where there is reduced necessity to retain the old knowledge, the model with CLTSQA-Framework could still obtain better performance. Table~\ref{Qualitative Results2} gives some examples of answers generated by ``FiD-Baseline'' and ``FiD-CLTSQA''.

\begin{table*}[t]
\caption{\label{Qualitative Results2} Examples of answers generated by ``FiD-Baseline'' and ``FiD-CLTSQA''.}
\begin{tabular*}{\linewidth}{l}
\toprule
\textbf{the most up-to-date data (evaluated on $\mathcal{D}_5^{test}$)}\\
\textbf{context}: He signed for South Coast Wolves after transferring from Sydney United ahead of the 2011 \\NSW Premier League season . Timpano left South Coast Wolves, signing for Dapto Dandaloo Fury \\ahead of their 2015 Illawarra Premier League campaign.\\
\textbf{question}: Which team did Jacob Timpano play for in 2013?\\
\textbf{FiD-Baseline $M_5$}: ``Sydney United'' \\
\textbf{FiD-CLTSQA $M_5$}: ``South Coast Wolves''\\
\textbf{label}: ``South Coast Wolves''\\
\midrule
\textbf{context}: Praveen Kumar was initially with the Royal Challengers Bangalore until 2010. In the \\Indian Premier League he played for Kings XI Punjab from 2011 to 2013.\\
\textbf{question}: Which team did Praveen Kumar play for in 2010?\\
\textbf{FiD-Baseline $M_5$}: `` '' (unanswerable)\\
\textbf{FiD-CLTSQA $M_5$}: ``Royal Challengers Bangalore''\\
\textbf{label}: ``Royal Challengers Bangalore''\\
\midrule
\textbf{context}: She was founder and chair of the Graduate Design Program at California College of the Arts \\( 2006–2012 ). \\
\textbf{question}: What was the name of the employer Brenda Laurel work for in 2012?\\
\textbf{FiD-Baseline $M_5$}: ``California College of the Arts''\\
\textbf{FiD-CLTSQA $M_5$}: `` '' (unanswerable)\\
\textbf{label}: `` '' (unanswerable)\\
\textbf{previous data (evaluated on $\mathcal{D}_1^{test}$)}  \\
\textbf{context}: It is known that Vytautas himself knew and spoke in the Lithuanian language with Jogaila. \\Struggle for power 1377–1384.\\
\textbf{question}: What was the residence of Vytautas in 1384?\\
\textbf{FiD-Baseline $M_1$}: `` '' (unanswerable)\\
\textbf{FiD-Baseline $M_5$}: ``Lithuania''\\
\textbf{FiD-CLTSQA $M_1$}: `` '' (unanswerable)\\
\textbf{FiD-CLTSQA $M_5$}: `` '' (unanswerable)\\
\textbf{label}: `` '' (unanswerable)\\
\midrule
\textbf{context}: University Hall , the first residential hall for women students in Scotland ,was founded at \\St Andrews University in 1895 ;Louisa Lumsden was appointed its first warden.\\
\textbf{question}: Which employer did Louisa Lumsden work for in 1895?\\
\textbf{FiD-Baseline $M_1$}: ``St Andrews University''\\
\textbf{FiD-Baseline $M_5$}: ``University Hall''\\
\textbf{FiD-CLTSQA $M_1$}: ``St Andrews University''\\
\textbf{FiD-CLTSQA $M_5$}: ``St Andrews University'' \\
\textbf{label}: ``St Andrews University''\\
\midrule
\textbf{context}: He was appointed Lord Advocate in 1775. His name appears in the 1776 minute book of the \\Poker Club. 2nd Earl of Shelburne and Pitt, he entered the cabinet in 1791 as Secretary of State for the \\Home Department.\\
\textbf{question}: Which position did Henry Dundas, 1st Viscount Melville hold in 1776?\\
\textbf{FiD-Baseline $M_1$}: ``Lord Advocate''\\
\textbf{FiD-Baseline $M_5$}: `` '' (unanswerable)\\
\textbf{FiD-CLTSQA $M_1$}: ``Lord Advocate''\\
\textbf{FiD-CLTSQA $M_5$}: ``Lord Advocate''\\
\textbf{label}: ``Lord Advocate''\\
\bottomrule
\end{tabular*}

\end{table*}

\subsection{Ablation Studies}

\subsubsection{The Contributions of TMR and TCL}

In order to further investigate the contributions of the two strategies brought by CLTSQA-Framework, we conduct ablation studies by building two more model variants upon ``FiD-CLTSQA'': 
\begin{itemize}
    \item \textbf{FiD-CLTSQA w/o TCL}, which only applies temporal memory replay
    \item \textbf{FiD-CLTSQA w/o TMR}, which only applies temporal contrastive learning.
\end{itemize}
Table~\ref{Ablation} shows the final evaluation result, where ``FiD-CLTSQA w/o TCL w/o TMR'' is indeed the baseline model ``FiD-Baseline''. 
The result answers our third research question: the temporal memory replay effectively alleviates forgetting of the previous knowledge, thus playing a more important role in the old subsets (``FiD-CLTSQA'' v.s., ``FiD-CLTSQA w/o TMR''). Differently, the temporal contrastive learning brings less significant but consistent performance improvement across all subsets (``FiD-CLTSQA'' v.s., ``FiD-CLTSQA w/o TCL'').
Overall, the CLTSQA-Framework benefits from both modifications.

\subsubsection{The Novelty of TMR}
In order to emphasize on the novelty of temporal memory replay, we conduct a comparative experiment by employing two more model variants upon ``FiD-Baseline'': 
\begin{itemize}
   \item \textbf{FiD-Baseline with MR}, which only applies memory replay which selects 10\% old knowledge from each previous subset and reuses them in the subsequent training process.
   \item \textbf{FiD-Baseline with TMR}, which only applies temporal memory replay demonstrated in section~\ref{sec:TMR}.
\end{itemize}
The experimental results shown in Table~\ref{Ablation2} provide compelling proof of the superiority of our temporal memory replay method over the memory replay.

\section{Conclusion}
In this study, we pioneered a novel task, Continual Learning for Temporal Sensitive Question Answering (CLTSQA). We first introduced a new dataset, CLTSQA-Data, to facilitate research in this area, followed by the introduction of a novel framework, CLTSQA-Framework, designed to assist models in handling temporally-sensitive QA in a continual learning context.
Our experimental results revealed that while the CLTSQA task poses fresh challenges for existing models, the proposed framework effectively equips the model to overcome these hurdles, resulting in improved performance.
We are confident that our contributions, encompassing both the dataset and the framework, will stimulate future research in this innovative direction. As we move forward, there is a need for further exploration of datasets and models to delve deeper into the complexities of CLTSQA.

\clearpage
\appendix


\section{CLTSQA-Data Statistics}
\label{sec:appendix1}

\begin{figure}[h]
\centering
\includegraphics[width=\columnwidth]{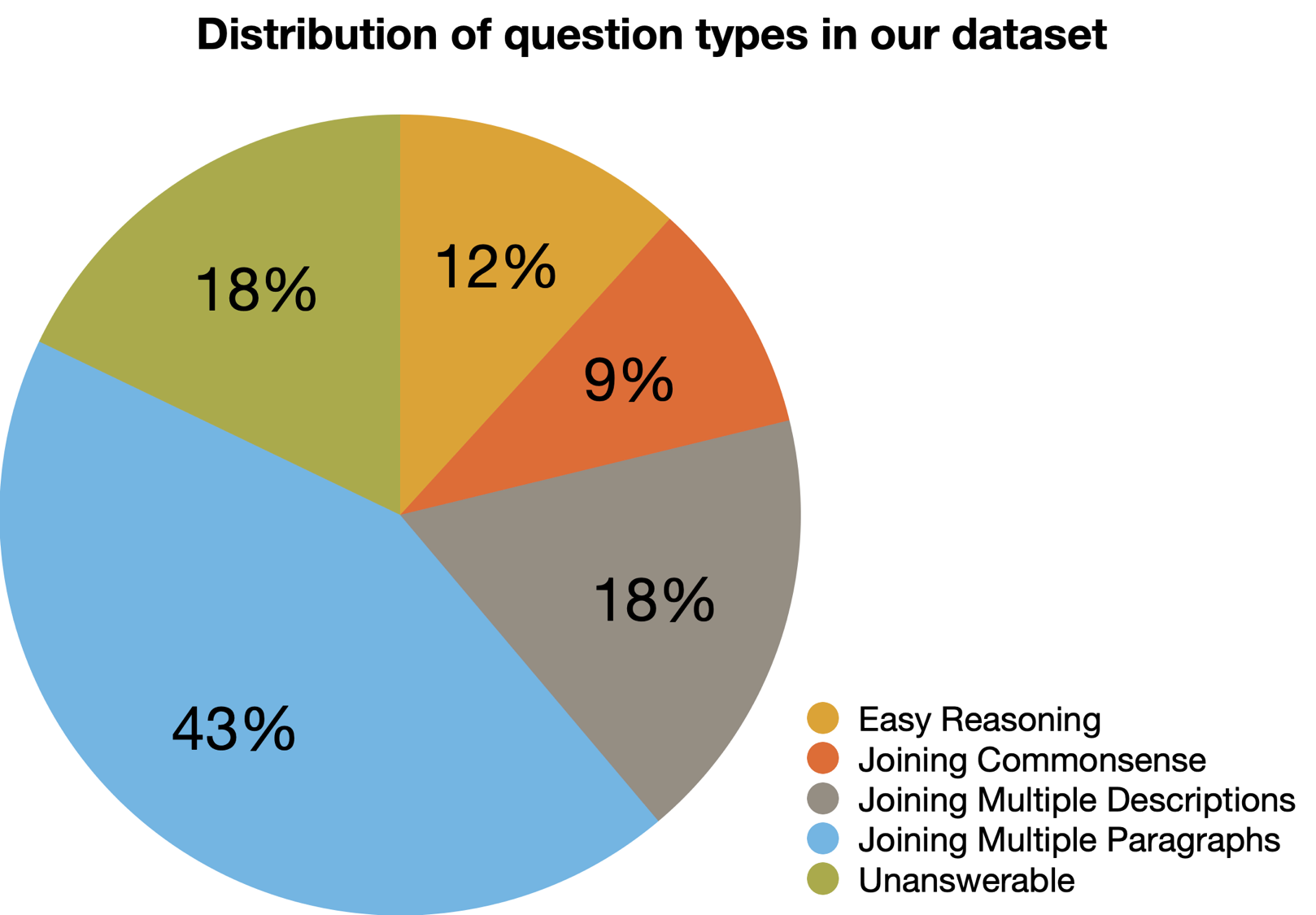}

\caption{Distribution of question types within the entire dataset.}
\label{distribution}

\end{figure}

\paragraph{Distribution of Question Types in CLTSQA-Data}
We investigated the various question types present in our dataset, which encompassed Easy Reasoning, Joining Commonsense, Joining Multiple Descriptions, Joining Multiple Paragraphs, and Unanswerable. Furthermore, we calculated the distribution of these question types within the entire dataset as Fig.~\ref{distribution} shows.

\paragraph{Examples in Question Types}
As shown in Table~\ref{Examples}, we present five different question types of our CLTSQA-Data, including context, question, and answer.

\section{Ablation Study on Temporal Memory Replay}
\label{sec:appendix5}
\begin{figure}[h]
\centering
\includegraphics[width=\columnwidth]{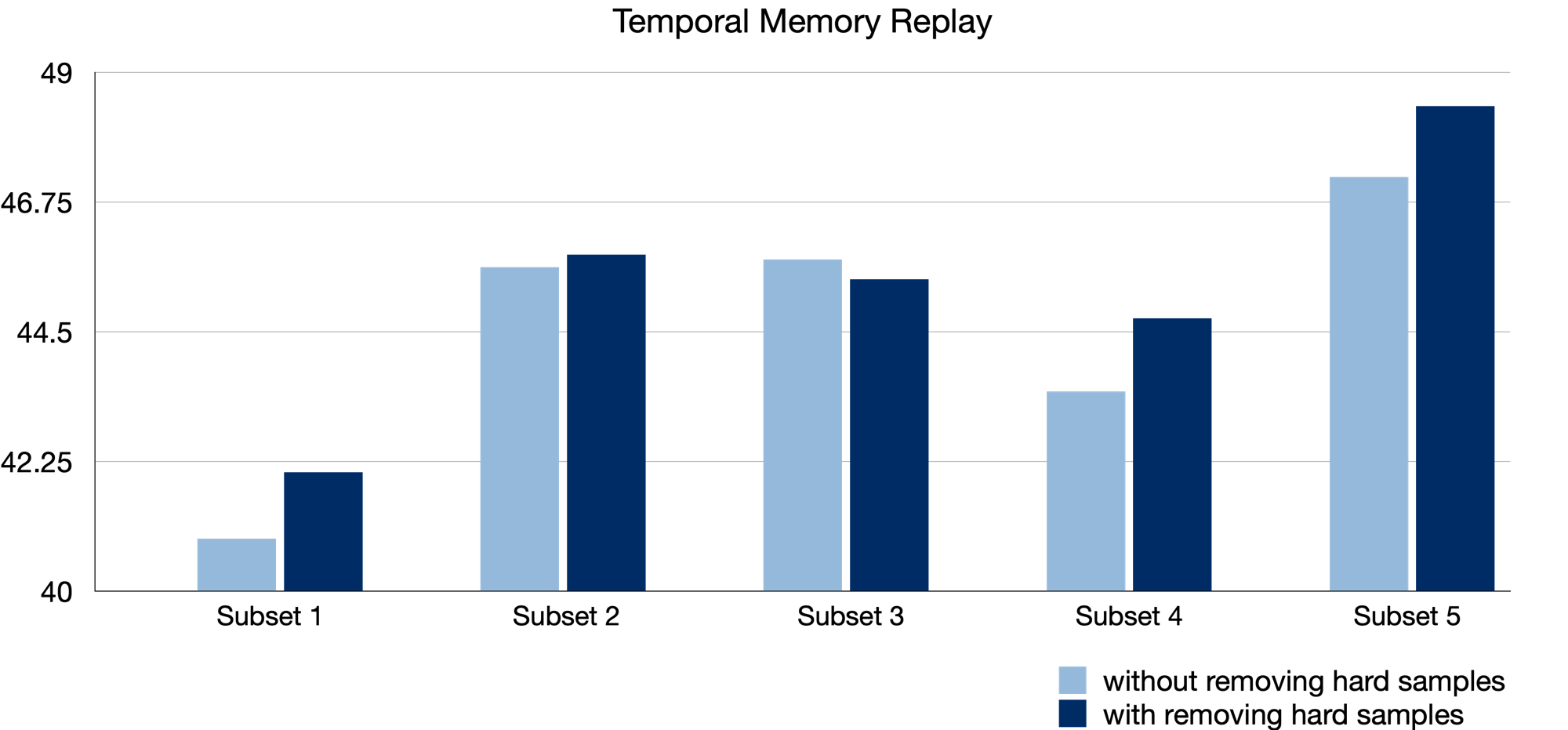}

\caption{Ablation study on temporal memory replay with / w.o the step of removing hard samples.}
\label{TMR}

\end{figure}

We investigated the performance of temporal memory replay with / w.o the step of removing hard samples, respectively. We assess model $M_5$ by $\mathcal{\overline{D}}_{5}^{dev}$. It can be seen from Fig.~\ref{TMR} that temporal memory replay with step removing hard samples has better performance.

\paragraph{Experimental Parameters}
\label{sec:appendix2}

The parameter settings for the two models, FiD and BigBird, used in the experiment are illustrated in Table~\ref{Parameters1} and Table~\ref{Parameters2}, respectively.

\begin{table}[h]
\caption{\label{Parameters1} FiD Model Parameters.}
\centering
\resizebox{0.8\columnwidth}{!}{
\begin{tabular}{l|c}
\toprule
\textbf{Parameters} & \textbf{FiD} \\
\midrule
max\_query\_length & 36 \\ 
max\_sequence\_length& 4096  \\
max\_answer\_length& 60   \\
max\_text\_length & 180  \\
learning\_rate &5$e^{-5}$ \\
adam\_epsilon &1$e^{-8}$    \\
pre\_gpu\_train\_batch\_size &1    \\
pre\_gpu\_eval\_batch\_size &1    \\
n\_gpu &4   \\
num\_train\_epochs &8   \\
\bottomrule
\end{tabular} }

\end{table}

\begin{table}[h]
\caption{\label{Parameters2} BigBird Model Parameters.}
\centering
\resizebox{0.8\columnwidth}{!}{
\begin{tabular}{l|c}
\toprule
\textbf{Parameters} & \textbf{BigBird} \\
\midrule
max\_query\_length & 36 \\ 
max\_sequence\_length& 3600  \\
doc\_stride& 2048   \\
learning\_rate & 5$e^{-5}$ \\
adam\_epsilon & 1$e^{-8}$    \\
pre\_gpu\_train\_batch\_size & 1    \\
pre\_gpu\_eval\_batch\_size & 1    \\
n\_gpu & 4   \\
num\_train\_epochs & 8   \\
\bottomrule
\end{tabular}
}
\end{table}

\begin{table*}[t]
\caption{\label{Examples} Examples of question types in CLTSQA-Data.}
\begin{tabularx}{\linewidth}{lX}
\toprule
\textbf{Easy Reasoning} & \\
\midrule
\textbf{context}: & ... Benedek Jávor, a proponent of the agreement, resigned from his position of parliamentary group leader, and Bernadett Szél were elected co-presidents of the LMP during the partys congress on 24 March 2013 ... \\
\textbf{question}: & Who was the head of LMP – Hungary's Green Party in 2013? \\
\textbf{label}: & ``Bernadett Szél'' \\
\midrule
\textbf{context}: & ... University Hall, the first residential hall for women students in Scotland, was founded at St Andrews University in 1895; Louisa Lumsden was appointed its first warden ... \\
\textbf{question}: & Which employer did Louisa Lumsden work for in 1895? \\
\textbf{label}: & ``St Andrews University'' \\
\midrule
\textbf{Joining Commonsense} & \\
\midrule
\textbf{context}: & He was purchased by the Kolkata Knight Riders at the 2011 IPL auctions for the next 3 years. \\
\textbf{question}: & Which team did the player Eoin Morgan belong to in 2012? \\
\textbf{label}: & ``Kolkata Knight Riders'' \\
\midrule
\textbf{context}: & He was Professor of Ancient History at the University of St Andrews from 1998 to 2014. \\
\textbf{question}: & Greg Woolf was an employee for whom in 2010? \\
\textbf{label}: & ``University of St Andrews'' \\
\midrule
\textbf{Joining Multiple Descriptions} & \\
\midrule
\textbf{context}: & In April 2014, Pohjanpalo renewed his contract with HJK, extending it to 2018. At the same time HJK extended his loan a further two years, which Pohjanpalo spent on loan at Fortuna Düsseldorf. \\
\textbf{question}: & Which team did Joel Pohjanpalo play for in 2015? \\
\textbf{label}: & ``Fortuna Düsseldorf'' \\
\midrule
\textbf{context}: & Tavares became CEO of Groupe PSA in 2014. Until January 16, 2021, he became the first chief executive officer of the multinational automobile group Stellantis. \\
\textbf{question}: & Carlos Tavares was an employee from whom in 2015? \\
\textbf{label}: & ``Groupe PSA'' \\
\midrule
\textbf{Joining Multiple Paragraphs} & \\
\midrule
\textbf{context}: & (paragraphs 1) ... He has been chair of the Labour Party in the House of Representatives since 10 June 2010 ... (paragraphs 2) On 20 February 2012, he resigned as leader of the Labour Party, ... \\
\textbf{question}: & What position did Job Cohen take in 2011? \\
\textbf{label}: & ``Leader of the Labour Party'' \\
\midrule
\textbf{context}: & (paragraphs 1) ... He was appointed Lord Advocate in 1775. His name appears in the 1776 minute book of the Poker Club... (paragraphs 2) 2nd Earl of Shelburne and Pitt, he entered the cabinet in 1791 as Secretary of State for the Home Department. \\
\textbf{question}: & Which position did Henry Dundas, 1st Viscount Melville hold in 1776? \\
\textbf{label}: & ``Lord Advocate'' \\
\midrule
\textbf{Unanswerable} & \\
\midrule
\textbf{context}: & ... In February 2016, she shared first place with Anastasia Bodnaruk and Soumya Swaminathan in the women's event of the Moscow Open, finishing third on tiebreak. In 2017 she competed again in the World Youth U16 Olympiad for Russia and her team won the gold medal ... \\
\textbf{question}: & Which title was conferred to Alexandra Obolentseva in 2017? \\
\textbf{label}: & `` '' \\
\midrule
\textbf{context}: & ... The P class were later re-allocated to shunting and station pilot duties. All eight locomotives passed into Southern Railway ownership at The Grouping in 1923 ... \\
\textbf{question}: & What operated SECR P class in 1921? \\
\textbf{label}: & `` '' \\
\bottomrule
\end{tabularx}
\end{table*}

\paragraph{Experimental Results}
\label{sec:appendix3}

Table~\ref{baseline}, \ref{Memory Replay}, \ref{Temporal Contrastive Learning}, \ref{CLTSQA Framework} show results of specific performance of each stage in FiD without CLTSQA-Framework, FiD with Temporal Memory Replay, FiD with Temporal Contrastive Learning and FiD with CLTSQA-Framework respectively. Each model $M_i$ is assessed by $\mathcal{\overline{D}}_{i}^{dev}$ and $\mathcal{\overline{D}}_{i}^{test}$.

\clearpage

\begin{table*}[t]
\caption{\label{baseline}
FiD-Baseline}
\centering
\resizebox{\linewidth}{!}{
\begin{tabular}{l|c|c|c|c|c|c|c|c|c|c|c|c|c|c|c|c|c|c|c|c}
\toprule
\multirow{3}*{ } & \multicolumn{4}{c|}{Subset1} & \multicolumn{4}{c|}{Subset2} & \multicolumn{4}{c|}{Subset3} & \multicolumn{4}{c|}{Subset4} & \multicolumn{4}{c}{Subset5}\\ 
\cline{2-21}
 & \multicolumn{2}{c|}{Dev} & \multicolumn{2}{c|}{Test} &  \multicolumn{2}{c|}{Dev} & \multicolumn{2}{c|}{Test}& \multicolumn{2}{c|}{Dev} & \multicolumn{2}{c|}{Test}& \multicolumn{2}{c|}{Dev} & \multicolumn{2}{c|}{Test}& \multicolumn{2}{c|}{Dev} & \multicolumn{2}{c}{Test} \\
\cline{2-21}
 &EM&F1&EM&F1&EM &F1&EM &F1&EM &F1&EM &F1&EM &F1&EM &F1&EM &F1&EM &F1\\
\midrule
{$M_1$} & 39.76 &50.66 & 33.54 & 46.04 \\
{$M_2$} & 39.18 & 49.52 & 35.33 & 45.10 & 45.69&55.26 & 44.94 & 55.57  \\
{$M_3$} & 37.39 & 47.77 & 32.10 & 42.99 & 45.27 & 55.46 & 43.31 & 54.77 & 46.01 & 55.46 & 48.13 & 56.71\\
{$M_4$} & 36.04 & 47.64 & 31.89 & 43.42 & 43.27 & 54.13 & 41.48 & 54.24 & 42.80 & 53.47 & 45.72 & 54.47 & 46.22 & 57.02 & 48.61& 58.62 \\
{$M_5$} & 34.25 & 45.29 & 29.97 & 40.22 & 43.84 & 53.89 & 39.26 & 50.95& 40.32 & 50.60 & 40.43& 49.48& 42.26 & 53.37 & 46.28& 56.60& 47.97 & 55.60 & 49.03& 56.29\\
\bottomrule
\end{tabular} 
}

\end{table*}

\begin{table*}[t]
\caption{\label{Memory Replay}
Temporal Memory Replay}
\centering
\resizebox{\linewidth}{!}{
\begin{tabular}{l|c|c|c|c|c|c|c|c|c|c|c|c|c|c|c|c|c|c|c|c}
\toprule
\multirow{3}*{ } & \multicolumn{4}{c|}{Subset1} & \multicolumn{4}{c|}{Subset2} & \multicolumn{4}{c|}{Subset3} & \multicolumn{4}{c|}{Subset4} & \multicolumn{4}{c}{Subset5}\\ 
\cline{2-21}
 & \multicolumn{2}{c|}{Dev} & \multicolumn{2}{c|}{Test} &  \multicolumn{2}{c|}{Dev} & \multicolumn{2}{c|}{Test}& \multicolumn{2}{c|}{Dev} & \multicolumn{2}{c|}{Test}& \multicolumn{2}{c|}{Dev} & \multicolumn{2}{c|}{Test}& \multicolumn{2}{c|}{Dev} & \multicolumn{2}{c}{Test} \\
\cline{2-21}
 &EM&F1&EM&F1&EM &F1&EM &F1&EM &F1&EM &F1&EM &F1&EM &F1&EM &F1&EM &F1\\
\midrule
{$M_1$} & 39.76 &50.66 & 33.54 & 46.04 \\
{$M_2$} & 40.72 & 50.92 & 35.33 & 46.42 & 44.41&53.68 & 45.85 & 56.06  \\
{$M_3$} & 40.08 & 50.30 & 35.74 & 47.16& 44.84 & 54.70 & 44.02 &55.11 & 44.68 & 54.81 & 46.72& 55.93\\
{$M_4$} & 40.91 & 50.38 & 37.66 &47.87 & 46.05 & 55.01 & 44.74 & 55.21 & 47.22 & 57.00 & 50.87 & 59.72 & 43.25 & 53.98 &49.57 & 59.35 \\
{$M_5$} & 42.06 & 52.44 &38.63 &49.12 & 45.84 & 55.63 &45.07 &55.49 & 45.41 & 55.68 &48.13 &57.22 & 44.73 & 54.81 &46.28 &56.93 & 48.42 & 55.72 &47.32 &54.81\\
\bottomrule
\end{tabular} 
}

\end{table*}

\begin{table*}[t]
\caption{\label{Temporal Contrastive Learning}
Temporal Contrastive Learning}
\centering
\resizebox{\linewidth}{!}{
\begin{tabular}{l|c|c|c|c|c|c|c|c|c|c|c|c|c|c|c|c|c|c|c|c}
\toprule
\multirow{3}*{ } & \multicolumn{4}{c|}{Subset1} & \multicolumn{4}{c|}{Subset2} & \multicolumn{4}{c|}{Subset3} & \multicolumn{4}{c|}{Subset4} & \multicolumn{4}{c}{Subset5}\\ 
\cline{2-21}
 & \multicolumn{2}{c|}{Dev} & \multicolumn{2}{c|}{Test} &  \multicolumn{2}{c|}{Dev} & \multicolumn{2}{c|}{Test}& \multicolumn{2}{c|}{Dev} & \multicolumn{2}{c|}{Test}& \multicolumn{2}{c|}{Dev} & \multicolumn{2}{c|}{Test}& \multicolumn{2}{c|}{Dev} & \multicolumn{2}{c}{Test} \\
\cline{2-21}
 &EM&F1&EM&F1&EM &F1&EM &F1&EM &F1&EM &F1&EM &F1&EM &F1&EM &F1&EM &F1\\
\midrule
{$M_1$} & 40.65 &51.21 & 36.22&47.59 \\
{$M_2$} & 33.29 & 43.64 & 30.03 &40.26  & 46.76&58.10 & 44.35 &55.79  \\
{$M_3$} & 28.62 & 39.09 & 24.74 & 34.47 & 47.12 & 58.07 & 44.09 & 55.96 & 48.09 & 57.73 & 49.93 & 58.57\\
{$M_4$} & 26.12 & 36.58 & 21.31 & 32.22 & 41.35 & 53.97 & 41.54 & 53.27 & 46.01 & 56.86 & 47.19 & 57.22 & 45.86& 56.61& 49.68&60.02  \\
{$M_5$} & 15.94 & 19.09 & 17.59 & 20.99& 17.86 & 21.53 &19.33 &23.43 & 17.95 & 22.82 & 18.94& 22.14& 42.83 & 52.65 &43.81 &53.40 & 48.93 & 56.91 &49.48 &57.40\\
\bottomrule
\end{tabular} 
}

\end{table*}

\begin{table*}[t]
\caption{\label{CLTSQA Framework}
FiD with CLTSQA-Framework}
\centering
\resizebox{\linewidth}{!}{
\begin{tabular}{l|c|c|c|c|c|c|c|c|c|c|c|c|c|c|c|c|c|c|c|c}
\toprule
\multirow{3}*{ } & \multicolumn{4}{c|}{Subset1} & \multicolumn{4}{c|}{Subset2} & \multicolumn{4}{c|}{Subset3} & \multicolumn{4}{c|}{Subset4} & \multicolumn{4}{c}{Subset5}\\ 
\cline{2-21}
 & \multicolumn{2}{c|}{Dev} & \multicolumn{2}{c|}{Test} &  \multicolumn{2}{c|}{Dev} & \multicolumn{2}{c|}{Test}& \multicolumn{2}{c|}{Dev} & \multicolumn{2}{c|}{Test}& \multicolumn{2}{c|}{Dev} & \multicolumn{2}{c|}{Test}& \multicolumn{2}{c|}{Dev} & \multicolumn{2}{c}{Test} \\
\cline{2-21}
 &EM&F1&EM&F1&EM &F1&EM &F1&EM &F1&EM &F1&EM &F1&EM &F1&EM &F1&EM &F1\\
\midrule
{$M_1$} & 40.65 &51.21 & 36.22&47.59 \\
{$M_2$} & 40.72 & 50.95 & 33.68 & 45.72 & 48.11 & 57.57 & 46.83 & 56.72 \\
{$M_3$} & 42.64 & 52.14 & 37.18 & 48.04 & 48.11 & 57.82 & 48.01 & 57.99 & 47.76 & 56.43 & 49.33 &57.72 \\
{$M_4$} & 44.43 & 53.85 & 36.08 &46.85 & 48.90 & 59.03 &47.03 & 58.81 & 50.77 & 59.83 & 50.28 & 59.68 & 46.72 & 57.17 &49.74 & 59.29  \\
{$M_5$} & 42.45 & 52.01 & 39.31 & 49.55& 47.97 & 57.95 &47.49 &57.76 & 47.96 & 57.05 &48.06 &56.84 & 46.08 & 55.95 & 49.12&59.16 & 49.71 & 57.48 &49.03 &57.06\\
\bottomrule
\end{tabular} 
}

\end{table*}
\end{document}